# Identification of Orchid Species Using Content-Based Flower Image Retrieval

Diah Harnoni Apriyanti[1]
[1]Purwodadi Botanic Garden
Indonesian Institute of Sciences (LIPI)
Pasuruan, Indonesia
diah007@lipi.go.id

Aniati Murni Arymurthy[2]
[2]Faculty of Computer Science
University of Indonesia
Depok, Indonesia
aniati@cs.ui.ac.id

Laksana Tri Handoko[3]
[3]Research Center for Informatics
Indonesian Institute of Sciences (LIPI)
Bandung, Indonesia
laks001@lipi.go.id

*Abstract*—In this paper, we developed the system for recognizing the orchid species by using the images of flower. We used MSRM (Maximal Similarity based on Region Merging) method for segmenting the flower object from the background and extracting the shape feature such as the distance from the edge to the centroid point of the flower, aspect ratio, roundness, moment invariant, fractal dimension and also extract color feature. We used HSV color feature with ignoring the V value. To retrieve the image, we used Support Vector Machine (SVM) method. Orchid is a unique flower. It has a part of flower called lip (labellum) that distinguishes it from other flowers even from other types of orchids. Thus, in this paper, we proposed to do feature extraction not only on flower region but also on lip (labellum) region. The result shows that our proposed method can increase the accuracy value of content based flower image retrieval for orchid species up to ± 14%. The most dominant feature is Centroid Contour Distance, Moment Invariant and HSV Color. The system accuracy is 85,33% in validation phase and 79,33% in testing phase.

*Keywords*— flower image retrieval; orchid; centroid contour distance; HSV color; Support Vector Machine

## I. INTRODUCTION

Manual identification of plant needs skill, more information, more thoroughness, and also more time. We depend so much on experts to find out what plant's name is. Right now, there is limited number of taxonomists in Indonesia [1].

Orchid is one of the biggest families in the class of flowers. Not only using leaf, stem, and root, but in general we can also use flower as the parameter in identifying orchid species. Indonesia is suspected to have high diversity of orchids as there are so many species of orchids that are not revealed yet in the world of science [2]. Identification of orchid is slightly different from previous research of plant identification [3,4,5,6,7,8]. It is because orchid has a unique part of flower called lip (labellum) that distinguishes it from other flowers even from other orchids [9,10] as shown in Fig. 1.

In this research, we propose semi automatic Content-based Image Retrieval (CBIR) of orchid species using MSRM (Maximal Similarity based on Region Merging) method for segmentation, shape and color feature for feature extraction, and SVM (Support Vector Macine) method for retrieving the images. We choose MSRM method for segmentation because MSRM method is easy to use and their segmentation is more accurate than another high level segmentation such as Graph Cut method [11]. Taxonomists usually use color and shape feature to manually distinguish one flower to another flower. The researchers used SVM in this research because it is proven to be very effective in former implementations [12,13,14].

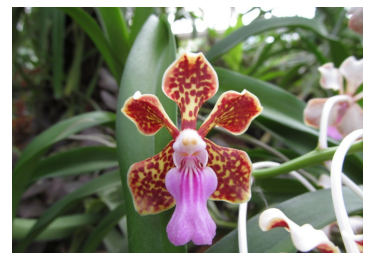

Fig. 1 Orchid flower

We analyzed the influence of CBIR orchid species system with lip of flower and without lip of flower. We conduct the analysis about the feature performance in order to find out the significant feature influencing the performance system. We also analyzed the performance system for validation phase and testing phase.

## II. MATERIAL AND METHOD

This section explains about the data and the proposed method including the pre processing, feature extraction, method of image retrieval and feature analysis.

### A. Data

We use 300 images of orchids, consist of 10 genus. Each genus consists of 3 species, and each species consist of 10 flower images taken from the front side. We obtain the data from various sources such as personal collection photos, colleague's photos, and photos from the internet with different size, resolution, distance shoot and light intensity. We used 5-cross validation for training phase to divide 300 images into 2 data sets, training sets and validations sets. We also use 30 images of orchids for testing. This 30 images are new images that have never been used for training/validation phase.

### B. Pre-Processing

Size of flower images are not more than 600x500 pixel in order to adjust with MSRM segmentation software [11] and the image resolution is 96 dpi (default result of imresize() function in Matlab). We did a little modification in the output of MSRM source code. If the MSRM output is white segmented object with black background, our output is color segmented object with black background. Segmentation was conducted twice, on flower region and lip region. After segmentation, we did morphological operation such as erosion, dilation, filling holes and 8-connected component to refine the image. The pre-processed images were ready for feature extraction process.

### C. Feature Extraction

We did the feature extraction on the flower region and lip region. The various data made us use the color and shape features that are invariant to scale, rotation, translation, light intensity, etc. Shape features that we used are distance from center to edge of flower/lip region, aspect ratio, roundness, moment invariant, and fractal dimension, while color features is HSV color with ignoring the V value. There are three features on behalf of distance from the center to the edge, that are CCD (Centroid Contour Distance), SF1, and SF2.

We chose CCD because it represents the flower shape by curve and is invariant to scale and rotation [4]. To be invariant to rotation, $0°$ start from the edge point which have the farthest distance from center of flower/lip region. In this research, we used edge point with multiple of $10°$[15]. CCD algorithm usually cannot run on convex shape (center point outer of the object), so we did a little modified algorithm. If there were no $10°$ or multiply, we use 0 for the distance. If there were $10°$ that cuts off two edge points, then the edge point that we use is the farthest.

According to [8], SF1 and SF2 can recognize flower well. We used SF1 to find out the sharpness of sepal and petal shape. We used SF2 to find out the pattern/shape of the flower based on average normalized distance. SF2 is invariant to scale. SF1 and SF2 can be computed by:

$$SF1 = \frac{R10}{R90} \quad (1)$$

$$SF_2 = \frac{1}{N}\sum_{i=1}^{N} D_i \quad (2)$$

where $R_{10}$ and $R_{90}$ are respectively the average distances among all di's which are smaller than $10^{th}$ percentile and larger than $90^{th}$ percentile of all di's:

$$d_i = \sqrt{(x_i - g_x)^2 + (y_i - g_y)^2}, 1 \leq i \leq N \quad (3)$$

$$g_x = \sum_{i=1}^{N} x_i \quad (4)$$

$$g_y = \sum_{i=1}^{N} y_i \quad (5)$$

N is the number of pixels on the flower boundary, $x_i$ and $y_i$ are respectively the x and y coordinates of the i-th boundary pixel. $D_i$ in (2) is the normalized distance defined as follows:

$$D_i = \begin{cases} 1 & d_i \geq R_{90} \\ \frac{d_i - R_{10}}{R_{90} - R_{10}}, & R_{10} < d_i < R_{90} \\ 0 & d_i \leq R_{10} \end{cases} \quad (6)$$

Aspect ratio is the ratio of physiological width and physiological length [16]. It is one of manual identification that is used by taxonomists. It is also invariant to scale, rotation and translation.

Roundness is taken as a feature because of its many variety of flower shapes so that each of orchid species may have different roundness value. Roundness can be computed by [8,16]

$$4\pi A/P^2 \quad (7)$$

where A is flower/lip area and P is perimeter of edge's flower/lip.

Moment Invariant (MI) was chosen because of its reliable capability as a shape feature. It was invariant to rotation, translation and dilatation. We used seven moment invariants deriving from second moment and third moment [17,18].

Fractal Dimension also has good performance for object recognition. Fractal Dimension can be computed as follows

$$D(s) = \frac{\log(N(s))}{\log(s)} \quad (8)$$

where N(s) is the number of boxes—which size is s—which filled of object information (pixel). D(s) is fractal dimension of object boxes—which size is s. We used dimension value of $4^{th}$, $5^{th}$, $6^{th}$, $7^{th}$ and the mean of its four dimension [18].

We chose HSV color because in this case, its performance is better than RGB color [19]. While the variation in illumination will greatly affect the recognition result, we used HSV color with the discard illumination (V) and then divided HS color space into 12x6 color [8] as shown in Fig. 2 and represented by $C_i$, $1 \leq i \leq 72$.

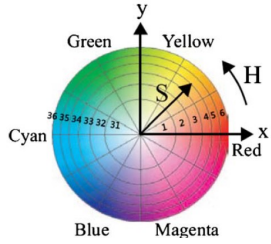

Fig. 2 HS color space divided into 12x6 color

The color coordinate of each cell can be represented by a pair of H & S values $(H_i, S_i)$, $1 \leq i \leq 72$. For each flower/lip region, a color histogram (notated CH(i), $1 \leq i \leq 72$), which describes the probability associated with each color cell $C_i$, will be generated. In [8], DC(1), DC(2) and DC(3) represent respectively the first three dominant color cells. In this research, because of the first dominant color cells is black (background color), we used second and third dominant color cells. Let $(dx_i, dy_i)$ and $p_i$ represent the coordinate vector and the corresponding probability of DC(i), $2 \leq i \leq 3$, where $dx_i = S_{DC(i)} \cos(H_{DC(i)})$ and $dy_i = S_{DC(i)} \sin(H_{DC(i)})$. Our color features can be summarized as follows

CF1: x-coordinate value of $DC_2$, $dx_2$

CF2: y-coordinate value of $DC_2$, $dy_2$

CF3: the probability of $DC_2$, $p_2$

CF4: x-coordinate value of $DC_3$, $dx_3$

CF5: y-coordinate value of $DC_3$, $dy_3$

CF6: the probability of $DC_3$, $p_3$

### D. Image Retrieval

To retrieve the image, we use SVM (Support Vector Machine) Multi Class method. SVM originate from Vapnik's statistical learning theory [20]. SVM is a kernel method. Its kernel function and its parameter is crucial in determining the performance [21]. The basic principle of SVM is to find an optimal separating hyperplane between the two data sets. First, the data was mapped into a high dimensional kernel space by using kernel function. The nearest data vector to the hyperplane is called support vector [22]. Optimal hyperplane can be found by maximizing the margin between the classes.

In this research, we used LibSVM library [23] with experiment on linear, polynomial, radial basis and sigmoid kernel. We changed the combination of parameter c from 1-100000, g from 0-32, r from 0-10 and d from 3-5 to get the best performance.

### E. Feature Analysis

To analyze the significant feature, we used feature selection by using Weka tool and also manually analyzing it by searching features with better performance than other features, then combined some dominant features as selected features to find out the performance system.

## III. RESULT AND DISCUSSION

From feature extraction on lip and flower region, we got 111 features that is shown in Tabel 1. SF1, SF2, Roundness and Aspect Ratio each have 1 feature, MI have 7 features, CCD have 36 features, Fractal Dimension have 5 features, and HSV color have 6 features.

Tabel 1 Result of Feature Extraction

| Feature | | Description |
|---|---|---|
| Flower Region | Lip of Flower Region | |
| f1 | f47 | SF1 |
| f2 | f48 | SF2 |
| f3 | f49 | Roundness |
| f4-f10 | f50-f56 | Moment Invariant |
| f11-f46 | f57-f92 | CCD |
| f93-f98 | f99-f104 | HSV Color |
| f110 | f111 | Aspect Ratio |
| f105-f109 | f105-f109 | Fractal Dimension |

As shown in Fig. 3, features that have individual good performance to recognize flower images, from the highest to the lowest, are CCD, HSV color, and Moment Invariant.

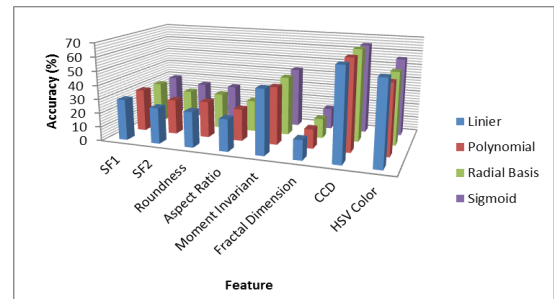

Fig 3. The individual ability of each feature to recognize flower image

From the result of Fig. 3, we combined features that have good performance. It was divided into several group. These combined features are CCD and MI (Group 1), CCD and HSV (Group 2), MI and HSV Color (Group 3), CCD, MI and HSV Color (Group 4). We also analyzed the performance system by using features that were used [8] in Group 5 (SF1, SF2, Roundness, HSV Color) and [18] in Group 6 (Moment Invariant and Fractal Dimension).

Result of feature selection using Weka was divided into 4 groups that shown in Table 2. Methods in Weka that we used are CFSubsetEval (using Best First and Genetic Search), WrapperSubsetEval (using Genetic Search), and PrincipalComponent (using Ranker).

Tabel 2 Result of Feature Selection using Weka Tool

| Attribute Subset Evaluator | Group | Search Method | Feature |
|---|---|---|---|
| CFS | 7 | Best First | f24, f25, f27, f31, f39, f48, f52, f75, f91, f93, f95, f98, f104, f107, f108 |
| | 8 | Genetic Search | f6, f14, f15, f17, f18, f22, f23, f24, f25, f26, f30, f31, f32, f37, f38, f39, f48, f52, f53, f55, f56, f73, f78, f80, f83, f84, f89, f91, f93, f95, f97, f98, f101, f104, f107, f108, f111 |
| Wrapper | 9 | Genetic Search | f9, f10, f14, f24, f28, f56, f60, f61, f62, f64, f65, f69, f76, f101, f102, f107 |
| Principal Component | 10 | Ranker | f1 until f59 |

According to [21] and after we experiment about the best kernel function which is suitable for this system, we chose the Radial Basis Kernel Function (c=30, g=0.009). It's because radial basis kernel with that parameter gives the best result in the training phase and is stable during the testing phase. Feature selection by Weka showed a little influence on the system accuracy, as shown in Fig. 4. The most powerful feature is using all of the features. It reaches 85,33 % accuracy. The second high accuracy is Group 4 (CCD, MI, and HSV color) with 82,67 % accuracy, and then Group 2 with 82 % accuracy.

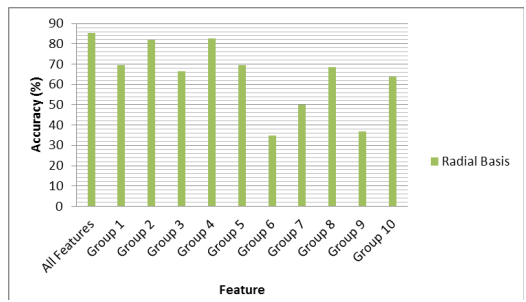

Fig. 4 System accuracy in various features combination

To find out the actual performance of our system, we tested 30 images that have never been trained before. Table 3 shows the accuracy recognition in validation phase and testing phase if we use only flower, only lip of flower and combine between flower and lip. It used 5-cross validation. From the table, we can see if we use not only the lip flower but also flower in the feature extraction, we get higher accuracy compared to using flower or lip of flower only.

Table 3 The system accuracy using SVM with flower, lip and combine flower+lip in Validation Phase and Testing Phase

| | Validation Phase | Testing Phase |
|---|---|---|
| **Flower** | 71 % | 66 % |
| **Lip** | 66 % | 56,67 % |
| **Flower+Lip** | 85,33 % | 79,33 % |

Table 4 The average time of feature extraction and recognition using flower, lip and combine flower+lip

| | Average Time |
|---|---|
| **Flower** | 1,4756 seconds/image |
| **Lip** | 1,3624 seconds/image |
| **Flower+Lip** | 2,7794 seconds/image |

Using the lip of flower and also the flower logically can increase the time of recognition. From Table 4, we can see that the computation time of using flower and lip increase twice rather than using only flower.

## IV. CONCLUSION

Identification of Orchid Species using Content Based Flower Image Retrieval System has been developed with SVM method. This system performs good on radial basis kernel (c=30, g=0.009) using all of features with 85,33% accuracy in validation phase and 79,33 % in testing phase. Using lip of flower on feature extraction can improve identification accuracy about ± 14 % and also increase the computation time. The average access time was 2,78 seconds per image. It is increased twice compare to using only flower. From this research, it also can show if the quote of "lip of flower orchid is unique" also done in this system. The feature analysis show if the dominant feature that influence significantly on this system is CCD, HSV Color and Moment Invariant.

ACKNOWLEDGMENT

We would like to thank Destario Metusala for sharing information about taxonomy of orchid, Mr. Suradi, Mr. Yoga from Bogor Botanic Garden, Mrs. Siti Nurfadilah, Mrs. Esti, Mrs. Nina from Purwodadi Botanic Garden, also Mr. Dede Setia Santoso from DD Orchid, Batu Malang for data support.